%% file: LBSVM.tex
\documentclass[runningheads]{llncs}
\usepackage{graphicx}
\usepackage{amsmath,amssymb}
\usepackage{color}
\usepackage[width=122mm,left=12mm,paperwidth=146mm,height=193mm,top=12mm,paperheight=217mm]{geometry}

\usepackage{booktabs}  

\usepackage{multirow}
\usepackage[ruled,linesnumbered]{algorithm2e}

\usepackage[table]{xcolor}
\usepackage{hhline}
\usepackage{rotating}
\usepackage{color}

\usepackage{subfigure}

\makeatletter
\newcommand*{\ccol}[1]{%
  \ifdim#1pt<.5pt\relax\else\color{white}\fi
  \edef\x{\noexpand\cellcolor[gray]{\strip@pt\dimexpr1pt-#1pt}}\x
  #1%
}
\newlength{\cellwidth}
\def\jline{\\\hhline{~*{10}{|-}|}}
\makeatother

\makeatletter 
  \newcommand\figcaption{\def\@captype{figure}\caption} 
  \newcommand\tabcaption{\def\@captype{table}\caption} 
\makeatother

\begin{document}
\pagestyle{headings}
\mainmatter

\title{Latent Bi-constraint SVM for Video-based Object Recognition}

\titlerunning{ }

\authorrunning{ }

\author{Yang Liu$^1$ \hspace{0.1cm} Minh Hoai$^2$ \hspace{0.1cm} Mang Shao$^1$ \hspace{0.1cm} Tae-Kyun Kim$^1$}
\institute{$^1$Imperial College London \hspace{0.5cm} $^2$Stony Brook University, SUNY}

\maketitle

\input{definitions}

\begin{abstract}
We address the task of recognizing objects from video input. This important problem is relatively unexplored, compared with image-based object recognition. To this end, we make the following contributions. First, we introduce two comprehensive datasets for video-based object recognition. Second, we propose Latent Bi-constraint SVM (LBSVM), a maximum-margin framework for video-based object recognition. LBSVM is based on Structured-Output SVM, but extends it to handle noisy video data and ensure consistency of the output decision throughout time. We apply LBSVM to recognize office objects and museum sculptures, and we demonstrate its benefits over image-based, set-based, and other video-based object recognition.
\end{abstract}

\section{Introduction}
Object recognition is an important research problem in computer vision with applications in a wide range of areas, including human-computer interaction, intelligent surveillance, industrial inspection, robotics, medical imaging. Because of its importance, object recognition has been extensively studied and many algorithms have been proposed. Most existing algorithms (e.g., \cite{boiman2008defense,bosch2007image,lazebnik2006beyond,FLi2005,DLowe,wang2010locality,yang2009linear,szegedy2015going}), however, are developed to recognize objects from images. They do not address the dynamics, clutter, and noisiness of video input. Only a few methods have considered videos, e.g., video-based descriptors~\cite{Brunet08,TLee,noceti2009spatio,JSivic2003,DStavens2010,DTa2009,yu2014innovative}, image-set matching~\cite{Arandjelovic05,Lee03,RWang2008,liu2014video,wan2015robust,wolf}, face recognition in video~\cite{cherniavsky2012semi,cinbis2011unsupervised,cour2009learning,nagendra2015video}, and video classification~\cite{karpathy2014large,donahue2015long}. However, these methods  are conceptually different from ours, which will be clarified in Sec.~\ref{related_work}.

The ability to recognize objects from video has many potential applications. 
Consider a concrete example of building a system that allows a museum's visitors to use their cell phones to recognize objects on display. 
In this situation, it is more beneficial and convenient to recognize objects from video input instead of images. 
First, many museum objects have 3D shape, and any image can only depict a single facet of an object. Thus, an image contains much less information than a video that provides multiple views of the object. Second, in a crowded museum environment, it is 
more convenient for a museum visitor to record a continuous video of an object, rather than to capture occlusion-free representative images of the object.

\begin{figure}[t]
\begin{center}
  \includegraphics[trim=0mm 5mm 0mm 0mm,width=0.79\textwidth]{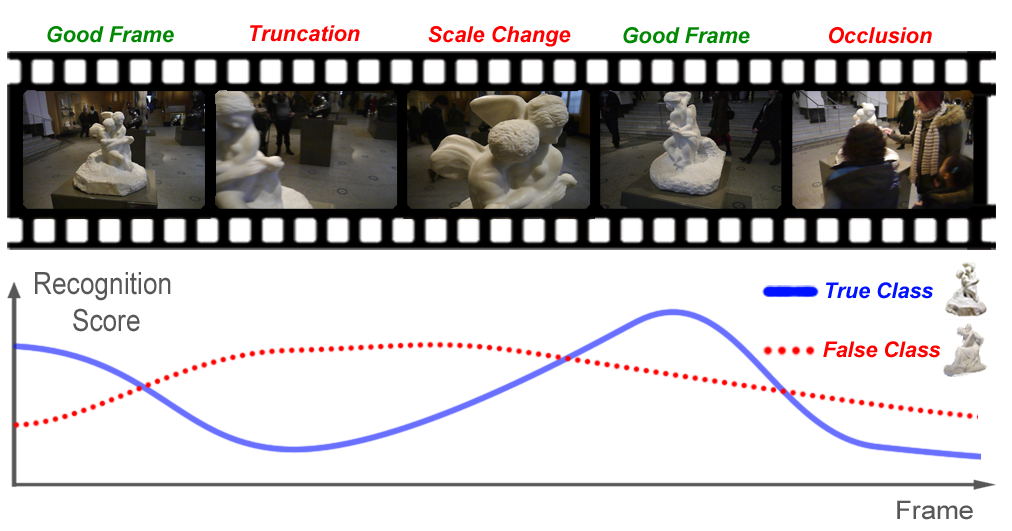}
\end{center}
\vskip -0.15in
\caption{{\bf Some challenges of video-based recognition.} 
The noise and variation of video data cause the recognition decision of frame-based approaches to fluctuate. The object is frequently recognized as the wrong class.}
\label{fig::intro}
\vspace{-0.1cm}
\end{figure}

Video-based object recognition, however, is challenging. 
Several highly important challenges are to: (1) handle the noisiness and variation of video data (e.g., not every video frame is occlusion free,
and videos can vary in length, object scale, and background clutter); (2) train classifiers when relatively few video examples of each object are present; (3) effectively use the entire video for recognition and avoid the fluctuation of the recognition decision over time. Some of these challenges are depicted in Fig.~\ref{fig::intro}. 

In this paper, we propose Latent Bi-constraint SVM (LBSVM), a novel algorithm for video-based object recognition. 
LBSVM is built on Structured-Output SVM~\cite{tsochantaridis2005large}, but extends it to address the challenges of recognizing objects from video input. 
LBSVM introduces two novel constraints and a latent variable. Its technical novelty is threefold: 1) The first constraint (Eq.~\ref{eqn:BSVM1}) expands the training video, associates the object label to all subsequences of each training video. This enforces all subsequences of training video to be correctly classified, enabling the recognition of an object from various view points. It also maximizes the usage of training data, reducing the need for a large number of training videos. 2) The second constraint (Eq.~\ref{eqn:BSVM2}) requires the monotonicity of the score function with respect to the inclusion relationship between subsequences of a video. This is to ensure the consistency of the recognition decisions.  3) The incorporation of the latent variable allows the monotonicity requirement to be satisfied, discarding bad views of an object due to such factors as occlusion and motion blur. The two constraints and the latent variable allow LBSVM to ground the recognition decision on the entire video, avoiding the inconsistency of the output decisions.

We will demonstrate the benefits of LBSVM for recognizing office objects and museum sculptures from videos recorded using a handheld camera. 
Videos of office objects were recorded in a cluttered office environment, while museum sculptures were recorded inside a crowded museum. These \textsl{in-the-wild} videos are challenging for object recognition due to various factors, including occlusion, background clutter, scale variation, illumination change, and motion blur. Note that the problem we tackle is video-level recognition, rather than frame-level recognition in a video. Hence, datasets with a large number of videos are required to evaluate the proposed method.

\section{Related work}
\label{related_work}
A majority of algorithms for object recognition~\cite{boiman2008defense,bosch2007image,lazebnik2006beyond,FLi2005,DLowe,wang2010locality,yang2009linear,szegedy2015going} assume the input is a single image. They can be adapted to work with video input by running image-based recognition on individual frames and subsequently accumulating the recognition scores~\cite{BLi2001,ren2009egocentric,ren2010figure,liuvideo15}. This approach, however, has several drawbacks. First, extracting frame-level descriptors and running image-based recognition for all individual frames are inefficient; this fails to consider the temporal similarity of nearby frames. Second, a simple approach for pooling evidence from all frames can lead to poor recognition performance due to dominance of irrelevant information from frames with occlusion or motion blur. Third, this approach fails to take into account the sequential nature of video data and may produce inconsistent decisions over time. 

Algorithms for object recognition in video exist. Most of them~\cite{Brunet08,TLee,JSivic2003,DStavens2010,DTa2009,yu2014innovative} propose to utilize the temporal information in video and improve local video descriptors by feature tracking. \cite{DStavens2010} tracks image patches using optical flow and learns an invariant feature for recognition. \cite{DTa2009} proposes an efficient search space for interest points to track features, which are then exploited to recognize objects. \cite{TLee} develops Best Template Descriptors (BTD) from video, quantizes them to generate a Bag-of-Words model, followed by a nearest neighbor classifier to recognize object in video. These methods improve feature descriptors for video, but they perform object recognition frame by frame. They neither address the insufficiency of training data nor ensure the consistency of frame recognition decisions.

A video is an ordered set of images. As such, video-to-video matching can be cast as set-to-set matching~\cite{Arandjelovic05,Lee03,RWang2008,liu2014video,wan2015robust,wolf}. \cite{wolf} proposes Kernel Principal Angles (KPA), which measures the intersection of two manifolds representing two sets. Since images in a video are collected continuously, they exhibit smooth data changes and can be well constrained on a low-dimensional manifold. KPA can be used to match video manifolds. However, image-set methods are often developed especially for face recognition~\cite{Arandjelovic05,Lee03,RWang2008}, including face recognition in videos~\cite{cherniavsky2012semi,nagendra2015video}, or character identification in television shows~\cite{cinbis2011unsupervised,cour2009learning}, which differ from the problem we tackle. A prerequisite of those works is face detection and tracking, but no detectors are available for generic objects in our case. Also, they assume temporal coherence cues, which might not hold for our videos. 

In this paper, we develop LBSVM to exploit the structured information contained in video for object recognition. LBSVM is based on SOSVM~\cite{tsochantaridis2005large}, which can learn a correlation function between a complex input space and a structured output space. SOSVM has been shown to be widely useful in many computer vision tasks, and it has also been extended in several ways. 
\cite{hare2011struck} uses SOSVM for adaptive tracking and detection. \cite{zhou2012learning} proposes two-layer SOSVM to recognize unsuccessful activities. \cite{hoai2012max} extends SOSVM for early event detection by anticipating the sequential nature of temporal events. \cite{shapovalova2012similarity} introduces similarity constraints for weakly supervised action classification, which performs classification and discriminative localization. \cite{wu2012view} utilizes kernelized SOSVM for recognizing human actions from arbitrary views, which implicitly infers the view label during both training and testing.

LBSVM learns and recognizes class labels of videos. It is different from \cite{Brunet08,TLee,JSivic2003,DStavens2010,DTa2009}, which perform frame-by-frame recognition in video. It is also different from previous works requiring finer-level annotation, such as~\cite{brostow2009semantic}, which labels the pixels in the first frame of a video and propagates the labels through the video.

\section{Latent Bi-constraint SVM}
\subsection{Learning formulation}

\begin{figure}[t]
\begin{center}
  \includegraphics[trim=0mm 0mm 0mm 0mm,height=4.2cm,width=0.6\textwidth]{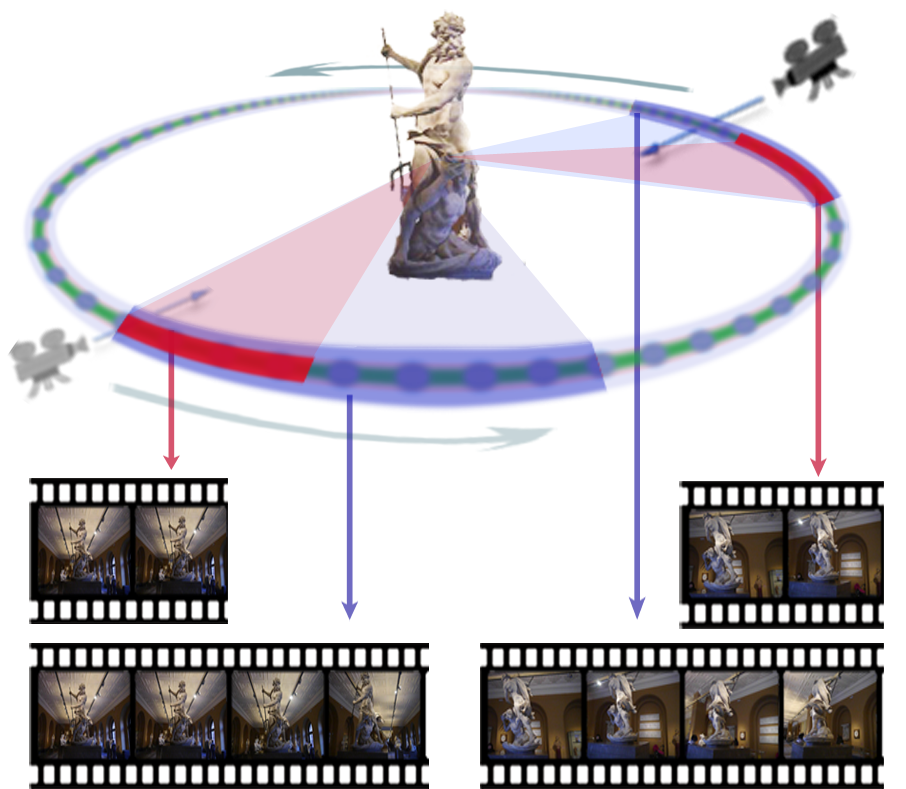}
\end{center}
\vskip -0.1in
\caption{Subsequences of a training video at different locations and scales.}
\label{fig::subsequences}
\end{figure}

\begin{figure}[t]
\centering
  \includegraphics[trim=30mm 0mm 0mm 15mm,height=5.5cm, width=0.8\textwidth]{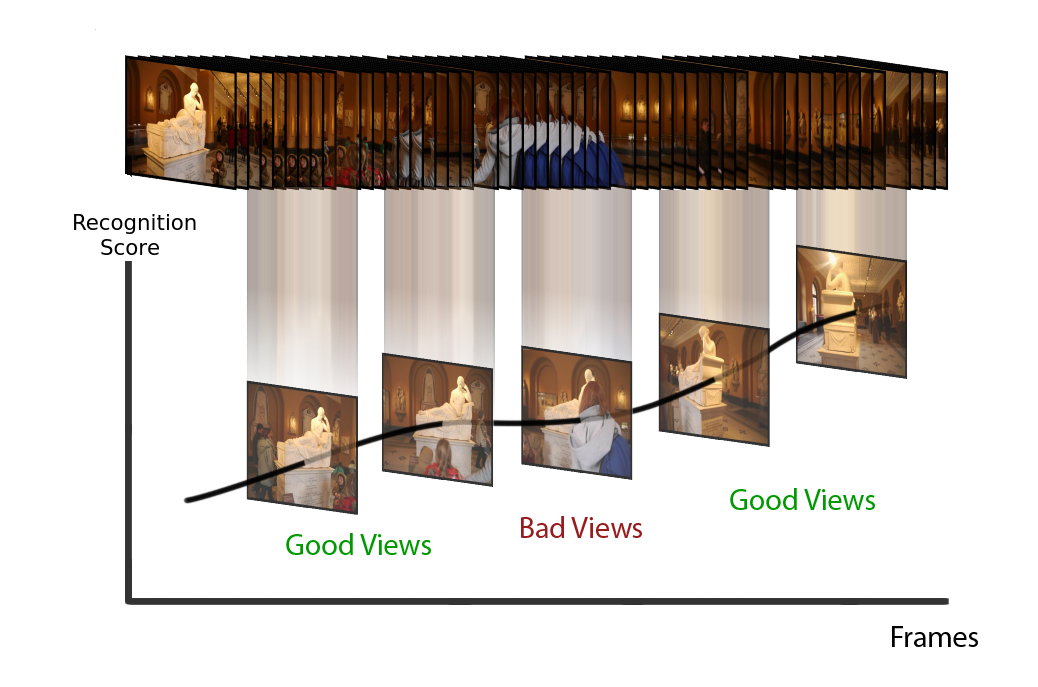}
  \vskip -0.1in
  \caption{Using a latent variable, LBSVM removes bad views to avoid the fluctuation of the recognition score, keeping it increasing.}
\label{fig::score graph}
\end{figure}

Let $\{\x_1, \cdots, \x_m\}$ be a set of training videos. Each video $\x_i$ depicts an object, and let $y_i$ be the label of that object. Let $\{\x_i^t\}$ be the set of all subsequences of video $\x_i$, at all locations and scales, as illustrated in Fig.~\ref{fig::subsequences}. We learn a LBSVM for video-based object recognition by solving the following optimization problem:
\begin{align}
\mymin{\w}\ &\frac{1}{2} ||\w||^2  \label{eqn:BSVM} \\
\myst & f_{\w}(\x_i^t, y_i) - f_{\w}(\x_i^t,y) \geq 1 \ \forall i, \forall t, \forall y \neq y_i, \label{eqn:BSVM1}\\
& f_{\w}(\x_i^t,y_i) - f_{\w}(\x_i^j, y_i) \geq \Delta(\x_i^t, \x_i^j) \label{eqn:BSVM2} \\
& \hspace{25ex} \forall i, \forall t, \forall j: \x_i^j \subset \x_i^t. \nonumber
\end{align}
Here $f_{\w}(\x,y)$ is the score function for a video segment $\x$ and a label $y$. We consider a linear recognition score function $f_{\w}(\x,y) = \w^T\psi(\x,y).$ $\psi(\x,y)$ is the joint feature mapping of the video segment $\x$ and the label $y$, and $\w$ is the parameter of the score function, which needs to be learned. Constraint~(\ref{eqn:BSVM1}) requires all subsequences of a training video to be correctly classified. This is based on the observation that if a video depicts an object then all subsequences of the video also depict the same object, possibly from different angles. This constraint essentially trains the system to recognize various views of the object. Constraint~(\ref{eqn:BSVM2}) requires monotonicity of the recognition function with respect to the inclusion relationship between video subsequences---the recognition score (confidence) of a video segment should not be lower than the recognition score of its subsequences. This constraint reflects a fact that a long video segment has more views of the object than a shorter one, and therefore, it should be recognized with higher confidence, as illustrated in Fig.~\ref{fig::score graph}. In Constraint~(\ref{eqn:BSVM2}), $\Delta(\cdot, \cdot)$ is the adaptive margin, and we use $\Delta(\x_i^t, \x_i^j) = 1 - \frac{|\x_i^j|}{|\x_i^t|}.$ By optimizing Eq.~(\ref{eqn:BSVM}), we obtain the function $f_{\w}$, that can be used for video-based object recognition. Given a new testing video $\x$, we predict the label of the object it depicts by finding the label that maximizes the score:
\begin{align}
y^* = \myargmax{y} f_{\w}(\x,y).
\end{align}
Here, $\x$ is the entire video, which contains more views of the object than any of its subsequences. 
As in the traditional formulation of SVM, the constraints are allowed to be violated by introducing slack variables:
\begin{align}
\mymin{\w,\alpha_i^t, \beta_i^t}\ &\frac{1}{2} ||\w||^2  + C_1 \sum_{i=1}^{m}\sum_t \alpha_i^t + C_2 \sum_{i=1}^{m}\sum_t \beta_i^t  \label{eqn:BSVM_full} \\
\myst & f_{\w}(\x_i^t, y_i) - f_{\w}(\x_i^t,y) \geq 1 - \alpha_i^t \ \forall i, \forall t, \forall y \neq y_i, \nonumber \\
& f_{\w}(\x_i^t,y_i) - f_{\w}(\x_i^j, y_i) \geq \Delta(\x_i^t, \x_i^j) - \beta_i^t  \nonumber \\
& \hspace{25ex} \forall i, \forall t, \forall j: \x_i^j \subset \x_i^t, \nonumber \\
& \alpha_i^t \geq 0, \beta_i^t \geq 0\ \forall i, \forall t. \nonumber
\end{align}

Although a video primarily depicts an object, the video might also contain not-so-informative frames due to several factors such as background clutter, occlusion, and motion blur. To filter out these irrelevant frames, we introduce latent variables into the feature mapping and the recognition function:
\begin{align}
f_{\w}(\x,y) = \max_{h} \w^T\psi(\x,y, h).
\label{eq::latent score function}
\end{align}
For a video sequence $\x$ that is represented by $l$ frames, $h$ is an indicator vector for selecting representative views of the object, and $\psi(\x,y,h)$ computes the feature representation on selected frames. As illustrated in~Fig.~\ref{fig::score graph}, where the use of the indicator $h$ is necessary to filter out bad views of the object to avoid the fluctuation of the recognition score. Details about the selector vector~$h$ and the feature representation will be described in Sec.~\ref{sec:feature}.

Once the model parameter $\w$ has been learned, inferring the label for a test video is done jointly with selecting good views of the object in the video:
\begin{align}
(y^*,h^*) = \myargmax{y,h} \w^T\psi(\x,y,h).
\end{align}

By exploiting the temporal similarity of nearby frames in a video, we can limit the domain of $h$ and only consider frames at a regular interval. This allows the above joint inference problem to be solved efficiently using exhaustive search. This will be described in more details in Sec.~\ref{sec:feature}.

\subsection{Optimization}
The constrained optimization  in Eq.~(\ref{eqn:BSVM_full}) is equivalent to the following unconstrained problem: 
\begin{equation} 
\mymins{\w} \frac{1}{2}{\left\| \w \right\|^2} + R(\w),
\label{eq::unconstrained}
\end{equation}
where
\begin{align*}
 R(\w) = \sum\limits_{i=1}^m \sum\limits_t \left(C_1\max \{0, R^1_{it} \} + C_2 \max \{0, R^2_{it} \} \right),
\end{align*}
and $R_1^{it}$ and $R_2^{it}$ are:
\begin{align*}
R^1_{it} & = \mathop{\max}\limits_{y \neq y_i}  {f_\w}(\x_i^t,y) + 1 -  {f_\w}(\x_i^t,{y_i}), \\
R^2_{it} & = \mathop {\max }\limits_{j: \x_i^j \subset \x_i^t} [{f_\w}(\x_i^j,{y_i}) + \Delta(\x_i^t,\x_i^j)] -  {f_\w}(\x_i^t,{y_i})  
\end{align*}

\begin{algorithm}[t]
\label{alg::learning}
\caption{LBSVM Learning}
\KwIn{$\{\x_i,y_i\}_{i=1}^{m}$: training videos and labels,
\hspace{0.001cm} $C_1$ and $C_2$: slack variable coefficients,
\hspace{0.001cm} $\epsilon$: gap threshold} 
\KwOut{Model parameter $\w$}
Initialize the model parameters $\w_1$ randomly

\For{$i \leftarrow 1 \hspace{0.12cm} \textbf{to} \hspace{0.12cm} m$}
	{
	Generate all subsequences ${\x_i^t}$ of $\x_i$
	}
\While {true}
{
	\For{$i \leftarrow 1 \hspace{0.12cm} \textbf{to} \hspace{0.12cm} m, each \hspace{0.12cm} t $}
	{
		- Fix $\w$, infer latent variables $({y^1_{it}},{h^1_{it}})$, $(x_i^{t'},h_{it}^{2})$, $h_{it}$ by  Eqs.~(\ref{eq::opt2}--\ref{eq::opt1})
		
		- Given updated $({y^1_{it}},{h^1_{it}})$, $(x_i^{t'},h_{it}^{2})$, $h_{it}$, recompute the feature representation for each video subsequence
	}
	Compute subgradient $c_{\w} = \frac{{\partial R(\w)}}{{\partial \w}}$ from Eq.~(\ref{eq::subgradient})\\
Update $\w$ by Eq. (13) of NRBM~\cite{do2009large}\\
	Compute $\w^*$ and $gap$ by Algorithm 1 of NRBM\\

	\If{$gap<\epsilon$}
        	  {	
         		  break;
    	 } 
} 
return $\w^*$\
\end{algorithm}

We use Non-convex Regularized Bundle Method (NRBM)~\cite{do2009large} to optimize for Eq.~(\ref{eq::unconstrained}). NRBM combines bundle methods and cutting plane techniques. It iteratively constructs an increasingly accurate piecewise quadratic lower bound of the objective function. In each iteration, a new cutting plane is found by the sub-gradient of the objective function and added to the piecewise quadratic lower bound approximation. The algorithm starts from a random $\w_1$ and generates a sequence of $\w_i$'s. The algorithm terminates when the 
 gap between the minimum of the approximation function and the value of the objective function is smaller than a predefined tolerant value. 

The sub-gradient of $R(\w)$ w.r.t. $\w$ can be computed from the gradients of $R^1$ and $R^2$. From the linear form in Eq.~(\ref{eq::latent score function}), the subgradient $\frac{{\partial R(\w)}}{{\partial \w}}$ is:
\begin{align}
  \sum\limits_{i=1}^m \sum\limits_{t}  &  \left\{C_1 [\psi (\x_i^t, y_{it}^{1}, h_{it}^{1}) - \psi (\x_i^t,{y_i},h_{it})] H(R^1_{it})  \right. \nonumber \\ 
+ & \left.  C_2  [\psi (\x_i^{t'},{y_i},h_{it}^2) - \psi (\x_i^t,{y_i},h_{it})] H(R^2_{it}) \right\} 
\label{eq::subgradient}
\end{align}
where $H(\cdot)$ is the Heaviside step function:
\[ H(R_{it}) = \left\{ 
  \begin{array}{l l}
    1 & \quad \text{if $R_{it} \ge 0$}\\
    0 & \quad \text{if $R_{it} < 0$},
  \end{array} \right.\]
and $({y^1_{it}},{h^1_{it}})$, $(x_i^{t'},h_{it}^{2})$, $h_{it}$ are inferred by:
\begin{align}
&({y_{it}^1},{h_{it}^1})  = \myargmax{y \neq y_i, \hspace{0.035cm} h} \w^T\psi(\x_i^t,y, h),\label{eq::opt2} \\
& (\x_i^{t'},h_{it}^2)   = \myargmax{j: \x_i^j \subset \x_i^t, \hspace{0.035cm} h} [\w^T\psi(\x_i^j,y_i, h)+\Delta(\x_i^t,\x_i^j)],  \label{eq::opt3} \\
& h_{it} = \myargmax{h} \w^T\psi(\x_i^t,y_i, h) \label{eq::opt1}.
\end{align}

In each iteration of NRBM, we infer $({y^1_{it}},{h^1_{it}})$, $(x_i^{t'},h_{it}^{2})$, $h_{it}$ and optimize model parameter $\w$ respectively:
\begin{enumerate}
\item Fix the model parameter $\w$, infer $({y^1_{it}},{h^1_{it}})$, $(x_i^{t'},h_{it}^{2})$, $h_{it}$ by Eqs.~(\ref{eq::opt2}--\ref{eq::opt1}). 

\item Fix $({y^1_{it}},{h^1_{it}})$, $(x_i^{t'},h_{it}^{2})$, $h_{it}$, finding a new cutting plane by Eq.~(\ref{eq::subgradient}) and add it to the quadratic piecewise approximation of NRBM, updating the model parameter $\w$ by minimizing the quadratic approximation.
\end{enumerate}

The learning process is shown in Algorithm~\ref{alg::learning}.

\section{Experiments}

This section introduces two datasets for video-based object recognition and demonstrates the benefits of LBSVM over frame-based, set-based, and video-based approaches. 

\subsection{Datasets}
\subsubsection{\small{Office Dataset}}  
\
\vspace{0.3cm}

\begin{figure*}[t]
\centering
  \includegraphics[trim=5mm 20mm 8mm 0mm, width=1\textwidth]{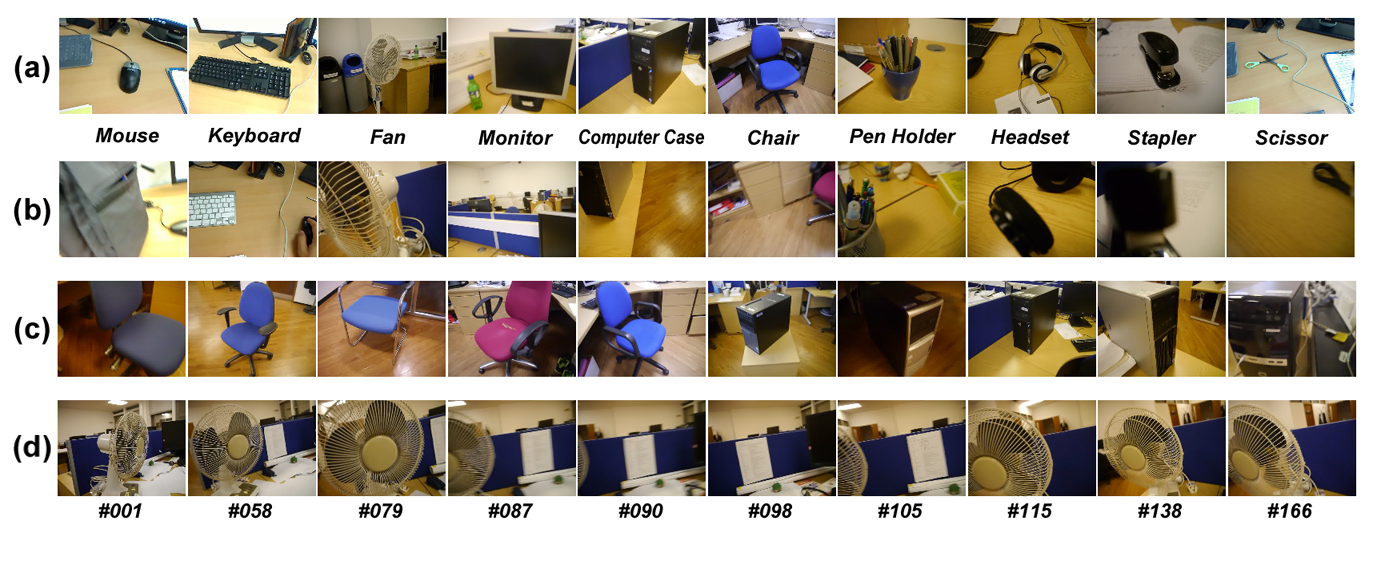}
  \caption{Example frames of \textsl{Office} dataset: (a) representative frames of the 10 object categories; (b) challenging frames of the 10 categories; (c) the 5 instances in the category of chair and computer case; (d) the variations of a fan in a video.}
  \label{fig::office}
\end{figure*}

\noindent The \textsl{Office} dataset contains 210 videos of 10 object categories in a cluttered office environment: mouse, keyboard, fan, monitor, computer case, chair, pen holder, headset, stapler, scissor. Some example frames are shown in Fig.~\ref{fig::office}(a). Each object category contains videos of 5 object instances (see Fig.~\ref{fig::office}(c)). Training data is a video spanning $360^{\circ}$ of one instance (Fig.~\ref{fig::subsequences}). Testing data is the remaining 20 videos of the other 4 instances, recorded with different variations (e.g., Fig.~\ref{fig::office}(d)). In total, there are 10 training videos and 200 testing videos. The durations of training videos are approximately 15 seconds, and the lengths of testing videos range from 6 to 10 seconds. 100 frames were extracted from each training video, and 20 frames per second were extracted from each testing video. The spatial resolution of all videos is $640\times480$ pixels. The \textsl{Office} dataset is challenging, with heavy clutters, extreme scales, illumination changes, and view shifting. In some frames, the object is even out of sight. Some challenging images are shown in Fig.~\ref{fig::office}(b). The variation of an object in a video is shown  in Fig.~\ref{fig::office}(d).

\subsubsection{\small{Museum Dataset}}
\ 
\vspace{0.3cm}

\begin{figure*}[t]
\centering
  \includegraphics[trim=5mm 14mm 8mm 15mm,width=1.0\textwidth]{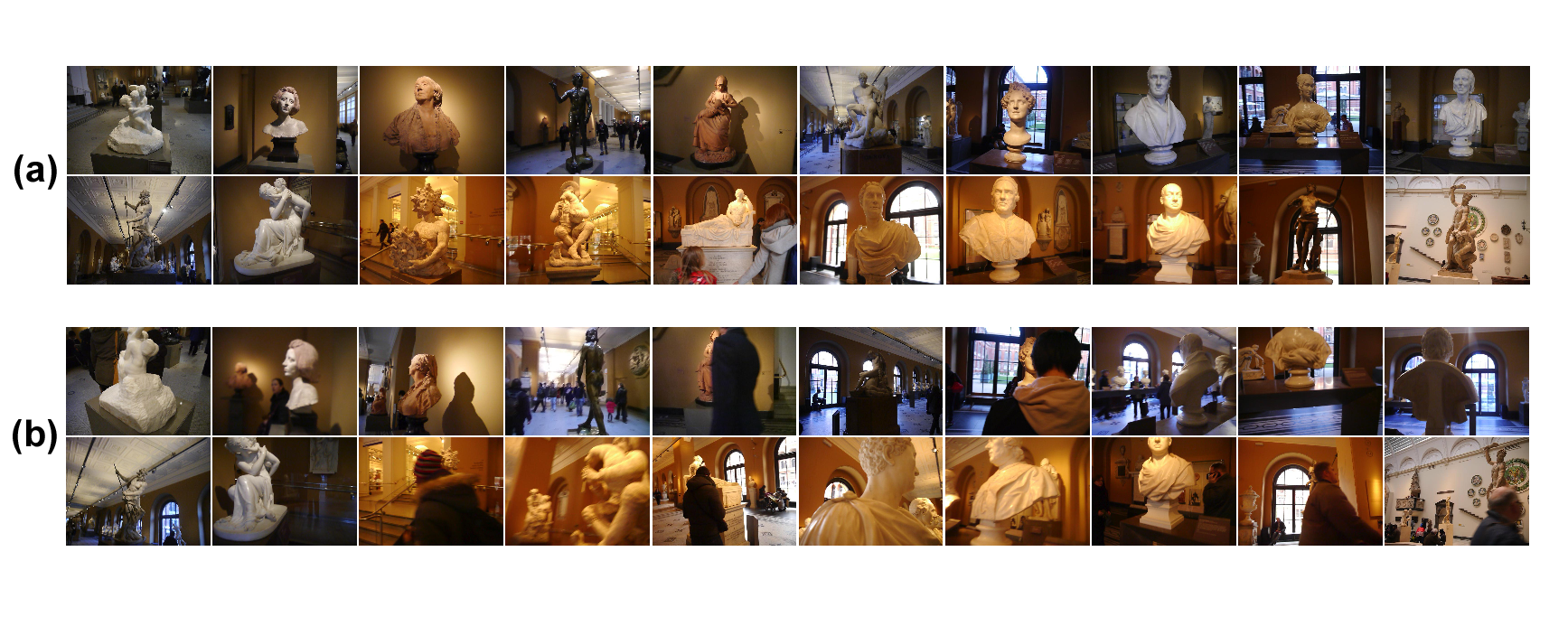}
  \caption{Example frames of \textsl{Museum} dataset: (a) representative frames of the 20 sculptures; (b) challenging frames of the 20 sculptures.}
  \label{fig::Dataset}
  \vspace{-0.4cm}
\end{figure*}

\noindent The \textsl{Museum} dataset contains 820 videos of 20 sculptures. The sculptures are 3D objects with low texture, and many of them have similar appearance. The sculptures includes portrait miniatures, statues, busts, as shown in Fig.~\ref{fig::Dataset}(a). Each sculpture has 41 videos: one is used for training and 40 for testing. The testing data is further divided into two equal and disjoint subsets, called Museum1 (easy) and Museum2 (hard). In total, there are 20 videos for training, 400 testing videos in Museum1 and the other 400 testing videos in Museum2. The videos in the training set, Museum1, Museum2 last for around 20 seconds, 6-10 seconds, and 5 seconds respectively. The videos of the Museum dataset have the same format as the \textsl{Office} dataset, including frame rate and spatial resolution. All videos were captured by a handheld camera. 

All videos were collected during rush hours, when the museum was crowded and the sculptures were surrounded by many people, the occlusion of the target sculpture and the background clutter were heavy. The training videos were taken  by moving the camera around each sculpture. The testing videos were collected by imitating the habits of average users. There are significant scale changes in the dataset. Some videos only partially cover the sculptures in the close distance, some other videos capture the sculptures as one tenth of the view. Most users are inexperienced photographer, the target sculpture is often not in the middle of the view, it sometimes slips out of the camera's point of view. Some challenging frames are shown in Fig.~\ref{fig::Dataset}(b).

\subsubsection{Feature representation} 
\label{sec:feature}
\

\vspace{0.3cm}
\noindent{\bf Feature extraction} For feature extraction, we use Dense SIFT (DSIFT)~\cite{DLowe,bosch2007image}. Subsequently, Spatial Pyramid Matching with Sparse Coding (ScSPM)~\cite{yang2009linear} and Bag-of-Words (BoW)~\cite{JSivic2003} were used as for aggregating descriptors for the \textsl{Office} dataset and the \textsl{Museum} dataset respectively. These features are common for all methods in the experiments. 

For ScSPM in \textsl{Office} dataset, all settings followed the standard way in~\cite{yang2009linear}. DSIFT was extracted from patches of $16\times16$ pixels from each sampled image with step size of 6 pixels. The codebook size was 1024, and we used spatial pyramid with 3 levels. The descriptors were aggregated using maximum pooling. The feature dimension was reduced to 150 by PCA.

For BoW in the \textsl{Museum} dataset, DSIFT was extracted at every 4 pixels with 4 patch sizes,  $16\times16$ pixels, $24\times24$ pixels, $32\times32$ pixels and $40\times40$ pixels. The codebook size was set as 300. The coded descriptors were aggregated by average pooling.

\noindent{\bf Subsequences} We generated subsequences for each training video as follows. First, for each video, we extracted 100 frames. The subsequences are sampled at 10 scales (from 10 to 100 frames) and at a regular interval (every second frame). In total, we generated 235 subsequences for each training video. These sequences correspond to multiple views of an object (Fig.~\ref{fig::subsequences}).

\noindent{\bf Subsequence representation} From each subsequence, $l$ frames were uniformly sampled. A subsequence is represented as a ScSPM or BoW feature vector, by pooling all quantized descriptors from the sampled frames. This could include the noise descriptors from the bad frames. To deal with this problem, we introduced a latent variable $h$, which selected half (empirically set, fixed in all experiments) of the $l$ frames, and encoded all possible selections as values of $h$. For each value of $h$, we computed a joint feature $\psi (\x,y,h)$ (ScSPM or BoW) by pooling the quantized descriptors from the selected frames, rather than from all the frames.

The temporal similarity of nearby frames in a video allows the subsequences to be subsampled. This limits the domain of the latent variable $h$ to be less than hundreds. Since a linear model is used in our algorithm, it is feasible to cope with all possible values of the latent variable.

\subsection{Compared methods}
We compared the proposed method with several frame-based, set-based, and video-based recognition methods. This section briefly describes these methods. 

\textbf{Average frame-based recognition (Avg-Frame)} This method takes a  frame, i.e., ScSPM~\cite{yang2009linear} or BoW~\cite{JSivic2003} feature vector, as the input of the classifier in both training and testing. A multi-class linear SVM is trained from the extracted frames of training videos. Each frame of testing videos is independently evaluated by the classifier. Finally the average frame recognition accuracy is reported. Here, we use LIBSVM~\cite{chang2011libsvm} with one-vs-one setting.  

\textbf{Accumulating frame recognition (Accum-Frame)} This method accumulates the frame-based recognition results (same as above) of all frames in a video to vote the video class~\cite{ren2009egocentric}. Three voting schemes are considered: Hard, Soft, and KNN voting. Hard voting uses the label results of the SVM. Soft voting adopts the probability estimation of the SVM. KNN voting selects 20 frames with the best SVM scores to vote for the video class. 

\textbf{Best Template Descriptor (BTD)} BTD \cite{TLee} learns video-based descriptors by feature tracking in training videos and uses a BoW model and a nearest neighbor classifier to recognize object in video frame by frame. Following~\cite{TLee}, we learn BTD descriptors from all subsequences of training videos, generating ScSPM (\textsl{Office}) or BoW (\textsl{Museum}) representations for them. A multi-class linear SVM classifier is trained from the subsequences, and frame-based recognition is performed for testing videos. Finally, soft voting is used to report the video-based recognition result. 

\textbf{Set-to-set matching (KPA)} Video-based object recognition can be solved by image-sets matching. We use Kernel Principal Angles (KPA)~\cite{liu2014video,wolf} to perform set-to-set matching between two videos (image sets). KPA takes two image sets as input, learns a manifold for each set, and compute the principal angles as the similarity between the two sets. Finally, a nearest neighbor classifier is used to classify a test video based on the similarity measurement. 

Detailed parameter settings are given in Secs.~\ref{sec::office} and~\ref{sec::museum_dataset}.

\subsection{Results on the Office dataset}
\label{sec::office}

\setlength{\tabcolsep}{8pt}
\begin{table}[t]
\vspace{0.15cm}
\begin{center}
\caption{Results on the \textsl{Office} and \textsl{Museum} datasets. The same feature representation is used on each dataset: ScSPM~\cite{yang2009linear} for \textsl{Office} and BoW~\cite{JSivic2003} for \textsl{Museum}. The proposed method LBSVM achieves the best accuracy on all three datasets.}
\vspace{0.15cm}
\label{fig:result}
\begin{tabular}{lrrr}
\toprule
 Algorithm & Office  & Museum1 & Museum2 \\
 \midrule
 Avg-Frame       & 65.1 & 67.3 & 56.7 \\
 Accum-Frame     & 74.5 & 91.5 & 73.5 \\
 BTD                     & 76.0 & 91.8 & 75.3 \\
 KPA                    & 79.5 & 95.0 & 85.8 \\
LBSVM (proposed) & \bf{84.5} & \textbf{98.8} & \textbf{91.5} \\
\bottomrule
\end{tabular}
\end{center}
\end{table}

\begin{figure}[t]
\begin{center}
  \includegraphics[trim=15mm 27mm 15mm 0mm, width=0.8\textwidth]{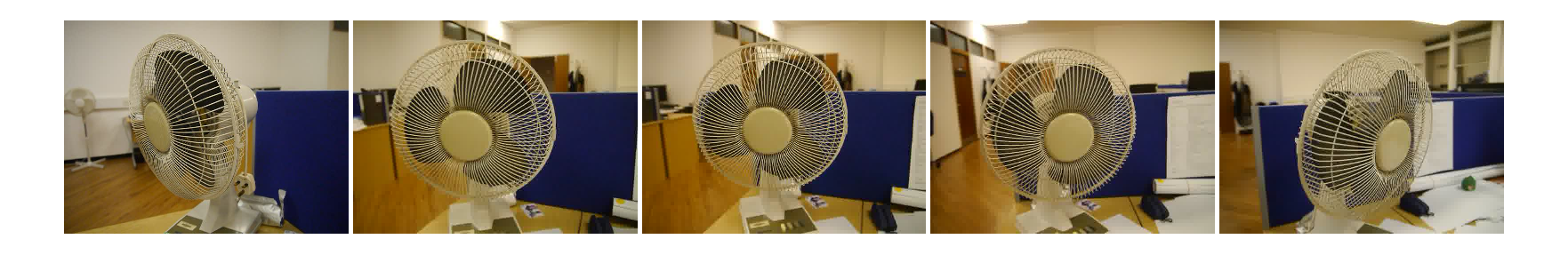}
\end{center}
\vskip -0.01in
\caption{The selected good views of fan by LBSVM.}
\label{fig::office_sel}
\vspace{-0.1cm}
\end{figure}

The second column of Tab.~\ref{fig:result}  shows the results of the various methods on the \textsl{Office} dataset. Avg-Frame with ScSPM, which is the state-of-the-art representation for image categorization, only achieves 65.1\% accuracy with 100 training images per category. For a comparison, ScSPM achieves 73.2 \% accuracy on the Caltech-101 dataset with only 30 training images per category~\cite{yang2009linear}. This demonstrates the challenges of the proposed \textsl{Office} dataset. The result of Accum-Frame in Tab.~\ref{fig:result} is by soft voting; Accum-Frame with hard voting and KNN voting achieve lower accuracies of 73.5\% and 72\%, respectively. All of these results are significantly better than the result of Avg-Frame. This indicates the importance of accumulating information in a video for object recognition. BTD is the video-based descriptor method using feature tracking. It slightly improves Accum-Frame. This is perhaps because the dataset was collected by a handheld camera producing short-range egocentric view, in which objects continuously shift or move out of sight, making feature tracking unstable.
KPA considers a video as a manifold and video recognition as manifold-to-manifold matching. This method yields better result than BTD and Accum-Frame. While BTD and Accum-Frame only rely on the available data, manifold can estimate the unseen data by interpolation and therefore has better generalization property. Both frame-based and aforementioned video-based recognition approaches, however, are inferior to LBSVM. The better accuracy of LBSVM can be credited to its ability to make use of all subsequence information from a video and at the same time it can filter out bad views of the object in the video by the latent variable. Fig.~\ref{fig::office_sel} displays the example of selected views by LBSVM.

The confusion matrices of BTD and LBSVM for recognizing objects from the \textsl{Office} dataset are shown in Fig.~\ref{fig::confusion matrix}. LBSVM outperforms BTD on most objects, yielding an accuracy of more than 80\% for all objects, except for \textsl{mouse} and \textsl{stapler}. This might be due to the low texture appearance of mouse and stapler objects.

\begin{figure}[t]
\vspace{0.2cm} 
    \subfigure[\hspace{-1.2cm}] 
    {
      \label{fig::confusion_matrix_BTD}
      {\tiny

\renewcommand{\arraystretch}{1.5} 
\setlength{\tabcolsep}{1.8pt}
\begin{tabular}{r|*{10}{>{\centering\arraybackslash}m{\cellwidth}}|}

\noalign{\gdef\w#1{\multicolumn{1}{c}{#1}}}
\cline{2-11} 
mouse         & \ccol{.30} & 0          & 0          & 0          & 0          & \ccol{.15} & 0          & \ccol{.35} & 0          & \ccol{.20} \\
keyboard      & \ccol{.10} & \ccol{.85} & 0          & 0          & 0          & \ccol{.05} & 0          & 0          & 0          & 0          \\     
fan           & 0          & \ccol{.15} & \ccol{.80} & \ccol{.05} & 0          & 0          & 0          & 0          & 0          & 0          \\      
monitor       & 0          & 0          & 0          & \ccol{1.0} & 0          & 0          & 0          & 0          & 0          & 0          \\
pc case       & 0          & 0          & 0          & \ccol{.10} & \ccol{.90} & 0          & 0          & 0          & 0          & 0          \\
chair         & 0          & 0          & 0          & 0          & \ccol{.10} & \ccol{.90} & 0          & 0          & 0          & 0          \\
pen holder    & 0          & \ccol{.25} & 0          & 0          & 0          & \ccol{.15} & \ccol{.60} & 0          & 0          & 0          \\
headset       & 0          & \ccol{.10} & 0          & 0          & 0          & 0          & 0          & \ccol{.80} & 0          & \ccol{.10} \\
stapler       & \ccol{.25} & 0          & 0          & 0          & 0          & 0          & 0          & 0          & \ccol{.55} & \ccol{.20} \\
scissor       & 0          & \ccol{.05} & \ccol{.05} & 0          & 0          & 0          & 0          & 0          & 0          & \ccol{.90} \jline

\w{} & \w{\begin{rotate}{-30}mouse\end{rotate}} & \w{\begin{rotate}{-30}keyboard\end{rotate}} & \w{\begin{rotate}{-30}fan\end{rotate}} & \w{\begin{rotate}{-30}monitor\end{rotate}} & \w{\begin{rotate}{-30}computer case\end{rotate}}  & \w{\begin{rotate}{-30}chair\end{rotate}} & \w{\begin{rotate}{-30}pen holder\end{rotate}} & \w{\begin{rotate}{-30}headset\end{rotate}} & \w{\begin{rotate}{-30}stapler\end{rotate}} & \w{\begin{rotate}{-30}scissor\end{rotate}}      \\
\w{} &\w{} &\w{} &\w{} &\w{} &\w{} &\w{} &\w{} &\w{} &\w{} &\w{} \\
\noalign{\global\let\w\undefined}
\end{tabular}
}
    }
    \subfigure[\hspace{-1.2cm}]
    {
      \label{fig::confusion_matrix_LBSVM}
      {\tiny

\renewcommand{\arraystretch}{1.5} 
\setlength{\tabcolsep}{1.8pt}
\begin{tabular}{r|*{10}{>{\centering\arraybackslash}m{\cellwidth}}|}

\noalign{\gdef\w#1{\multicolumn{1}{c}{#1}}}
\cline{2-11} 
mouse         &\ccol{.55}& 0          & 0          & 0          & 0          & \ccol{.05} & 0          & \ccol{.30} & 0          & \ccol{.10} \\
keyboard      & \ccol{.15} & \ccol{.80} & 0          & 0          & 0          & \ccol{.05} & 0          & 0          & 0          & 0          \\     
fan           & 0          & \ccol{.05} & \ccol{.85} & \ccol{.10} & 0          & 0          & 0          & 0          & 0          & 0          \\      
monitor        & 0          & 0          & 0          & \ccol{1.0} & 0          & 0          & 0          & 0          & 0          & 0          \\
pc case & 0          & 0          & 0          & \ccol{.15} & \ccol{.85} & 0          & 0          & 0          & 0          & 0          \\
chair         & 0          & 0          & 0          & 0          & \ccol{.05} & \ccol{.95} & 0          & 0          & 0          & 0          \\
pen holder & 0          & 0          & 0          & \ccol{.05} & \ccol{.05} & 0          & \ccol{.90} & 0          & 0          & 0          \\
headset       & \ccol{.05} & 0          & 0          & 0          & \ccol{.05} & 0          & 0          & \ccol{.85} & 0          & \ccol{.05} \\
stapler       & \ccol{.10} & 0          & 0          & 0          & 0          & \ccol{.10} & \ccol{.10} & 0          & \ccol{.70} & 0          \\
scissor       & 0          & 0          & 0          & 0          & 0          & 0          & 0          & 0          & 0          & \ccol{1.0} \jline

\w{} & \w{\begin{rotate}{-30}mouse\end{rotate}} & \w{\begin{rotate}{-30}keyboard\end{rotate}} & \w{\begin{rotate}{-30}fan\end{rotate}} & \w{\begin{rotate}{-30}monitor\end{rotate}} & \w{\begin{rotate}{-30}computer case\end{rotate}}  & \w{\begin{rotate}{-30}chair\end{rotate}} & \w{\begin{rotate}{-30}pen holder\end{rotate}} & \w{\begin{rotate}{-30}headset\end{rotate}} & \w{\begin{rotate}{-30}stapler\end{rotate}} & \w{\begin{rotate}{-30}scissor\end{rotate}}   \\
\w{} &\w{} &\w{} &\w{} &\w{} &\w{} &\w{} &\w{} &\w{} &\w{} &\w{} \\
\noalign{\global\let\w\undefined}
\end{tabular}

}
    }
  \vspace{-0.2cm}
  \caption{Confusion matrices for recognition on the \textsl{Office} dataset: \subref{fig::confusion_matrix_BTD} BTD; \subref{fig::confusion_matrix_LBSVM} LBSVM.}
  \label{fig::confusion matrix}
  \vspace{0.2cm}
 \end{figure}

\begin{figure}[t]
\begin{center}
  \includegraphics[trim=20mm 95mm 20mm 90mm,width=0.78\textwidth, height = 5.8cm]{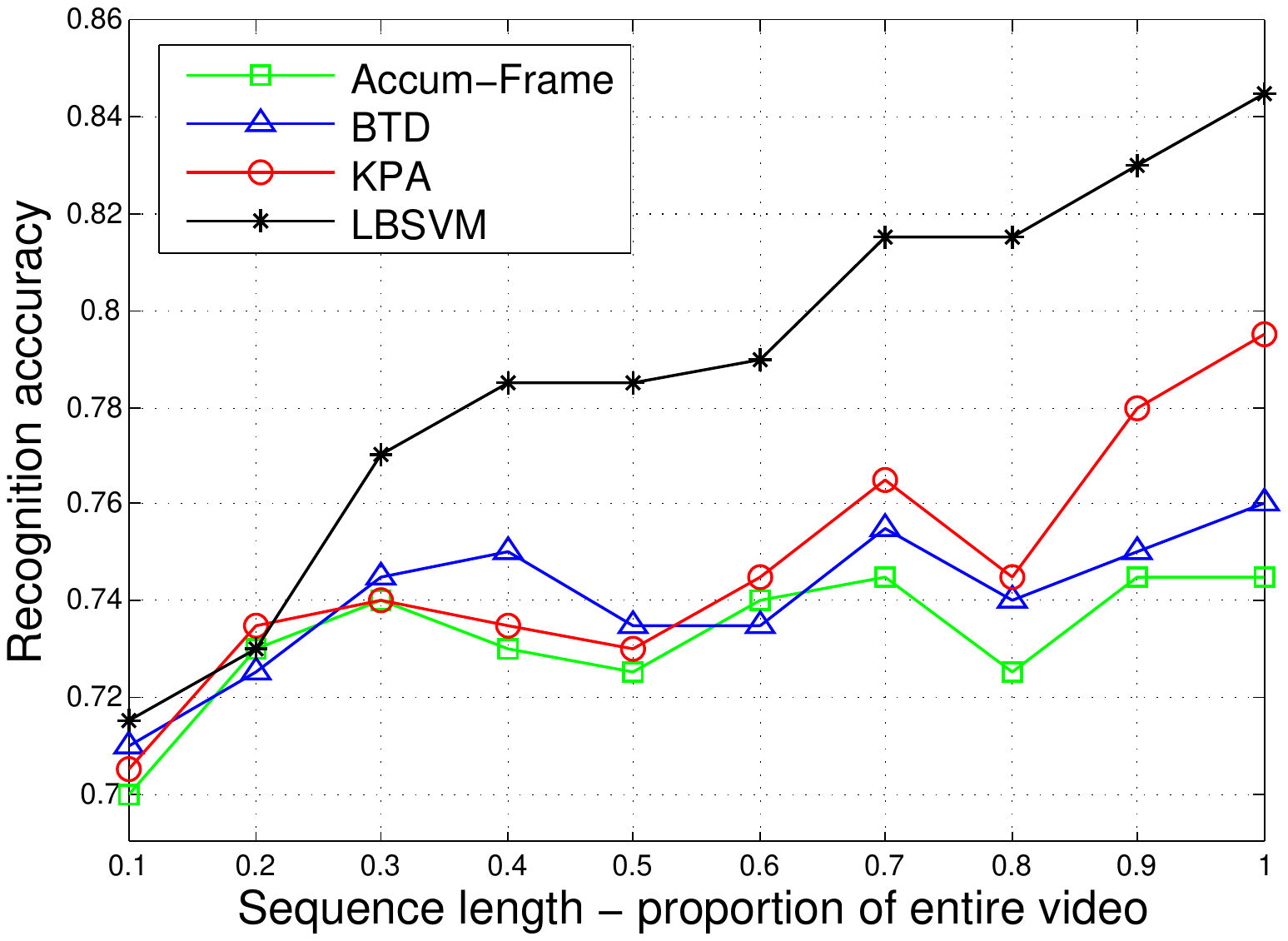}
\end{center}
\caption{The proposed LBSVM can accumulate information effectively and keep the recognition accuracy increasing monotonically, while the results of Accum-Frame, BTD, and KPA fluctuate.}
\label{fig::experimental monotonicity}
\end{figure}

To analyze the consistency of the recognition decision, we evaluate the recognition accuracy over time, by running the recognition algorithm on subsequences of testing videos. Fig.~\ref{fig::experimental monotonicity} plots the recognition accuracy against the length of video subsequence (from 10\% to 100\% of testing videos). As the sequences become longer and more views appear, the results of Accum-Frame, BTD and KPA methods fluctuate, while the proposed method can accumulate information effectively and keep the recognition accuracy increasing. Especially in some intervals, where the compared methods decrease dramatically, the recognition performance of the proposed method still increases or remain the same. Hence, LBSVM can ground the recognition on the entire video.

Tab.~\ref{fig:result_alt} reports the performance of two variants of  LBSVM. BSVM is LBSVM without the ability to discard uninformative frames. As can be seen, it does not perform as well as LBSVM. 
This emphasizes the importance of latent variable and view selection. SCSVM is BSVM without the monotonicity constraint (Constraint~(\ref{eqn:BSVM2})), and it has even lower recognition accuracy.

\setlength{\tabcolsep}{5pt}
\begin{table}[t]
\vspace{0.15cm}
\begin{center}
\caption{\textbf{Comparison with variant methods}. BSVM is LBSVM without latent variables. SCSVM is BSVM without enforcing monotonicity constraint (Constraint~(\ref{eqn:BSVM2}))}
\label{fig:result_alt}
\begin{tabular}{lrrr}
\toprule
 Algorithm & Office  & Museum1 & Museum2 \\
 \midrule
 SCSVM & 80.0 & 94.5 & 86.0 \\
 BSVM & 82.0 & 96.8 & 89.3 \\
LBSVM (proposed) & \bf{84.5} & \textbf{98.8} & \textbf{91.5} \\
\bottomrule
\end{tabular}
\end{center}
\vspace{-0.35cm}
\end{table}

LBSVM is efficient. In training, using PCA for reducing the dimension of feature vectors, it took about 2 hours for \textsl{Office} dataset with the maximum iteration of 300. In testing, it took 11ms to classify a video. This excludes the time for feature extraction, which is common for all methods. Since LBSVM needs to compute features in fewer sampled frames, the time for feature extraction is also largely reduced. These timing figures were measured on an Intel Core i7 3.4GHZ$\times$8 processor with 8GB RAM, for a Matlab implementation of LBSVM.

The parameters of the compared methods were set to report the best accuracies. Specifically, in KPA method, the Gaussian kernel was used with the bandwidth parameter $\gamma = 1$, and using the first principal angle as similarity got the best result. In SCSVM, the slack variable coefficient was set as $C=1\times10^{-4}$. In BSVM, the slack variable coefficients were $C_1=C_2=0.5\times10^{-4}$. For LBSVM, the slack variable coefficients were set the same as those in BSVM. The number of sampled frames for view selection was $l=10$. The stopping criterion for SCSVM, BSVM and LBSVM was $\epsilon = 0.01$.

\subsection{Results on Museum dataset}
\label{sec::museum_dataset}

The results of LBSVM and several other methods on the \textsl{Museum} dataset are shown in the last two columns of Tab.~\ref{fig:result}. The low results of Avg-Frame show the challenges of the two \textsl{Museum} datasets. Accum-Frame (using soft-voting) significantly outperforms Avg-Frame.  The other voting schemes, hard-voting and KNN-voting, do not perform as well, achieving 90.0\% and 86.5\% on Museum1, and 71.5\% and 71.5\% on Museum 2. BTD, based on video descriptors, is slightly better than Accum-Frame. KPA (manifold matching) and SCSVM (learning on all subsequences) achieved comparable performance, but were outperformed by BSVM (ensuring the consistency of recognition decision). The proposed method LBSVM, by view selection and information accumulation, yielded the best results on both datasets.

The Gaussian kernel of KPA method had the bandwidth parameter $\gamma = 0.9$. The slack variable coefficients for SCSVM and LBSVM (and also BSVM) were set as $C=10^{-5}$ and $C_1=C_2=0.5\times10^{-5}$. Other parameters were the same as \textsl{Office} dataset in Section~\ref{sec::office}.

\section{Conclusions}

We proposed two new datasets and a novel algorithm, LBSVM, for video-based object recognition. 
LBSVM is based on Structured-Output SVM, but extends it to handle noisy video data and ensure consistency of the output decisions. LBSVM introduces two novel constraints. The first constraint 
expands training videos and requires all subsequences to be correctly classified, training the classifier to recognize testing videos of various views. The second constraint  imposes monotonicity of the score function with respect to the inclusion relationship between subsequences of a video. Furthermore,
LBSVM incorporates latent variables for view selection, filtering out bad views of an object in a video. The latent variable, together with the two novel constraints, allow LBSVM to ground the recognition decision on the entire video, avoiding the inconsistency of the output decisions.
In training, we optimized the parameters of an LBSVM and the latent variables iteratively. In testing, we jointly  inferred the latent variable and the class label to maximize the score function. We showed that our algorithm outperformed frame-based, set-based, and other video-based object recognition approaches on the two new datasets for video-based object recognition.

\bibliographystyle{splncs}
\bibliography{egbib}
\end{document}

%% file: definitions.tex
\def\mA{\mathcal{A}}
\def\mB{\mathcal{B}}
\def\mC{\mathcal{C}}
\def\mD{\mathcal{D}}
\def\mE{\mathcal{E}}
\def\mF{\mathcal{F}}
\def\mG{\mathcal{G}}
\def\mH{\mathcal{H}}
\def\mI{\mathcal{I}}
\def\mJ{\mathcal{J}}
\def\mK{\mathcal{K}}
\def\mL{\mathcal{L}}
\def\mM{\mathcal{M}}
\def\mN{\mathcal{N}}
\def\mO{\mathcal{O}}
\def\mP{\mathcal{P}}
\def\mQ{\mathcal{Q}}
\def\mR{\mathcal{R}}
\def\mS{\mathcal{S}}
\def\mT{\mathcal{T}}
\def\mU{\mathcal{U}}
\def\mV{\mathcal{V}}
\def\mW{\mathcal{W}}
\def\mX{\mathcal{X}}
\def\mY{\mathcal{Y}}
\def\mZ{\mathcal{Z}}

\def\1n{\mathbf{1}_n}
\def\0{\mathbf{0}}
\def\1{\mathbf{1}}

\def\A{{\bf A}}
\def\B{{\bf B}}
\def\C{{\bf C}}
\def\D{{\bf D}}
\def\E{{\bf E}}
\def\F{{\bf F}}
\def\G{{\bf G}}
\def\H{{\bf H}}
\def\I{{\bf I}}
\def\J{{\bf J}}
\def\K{{\bf K}}
\def\L{{\bf L}}
\def\M{{\bf M}}
\def\N{{\bf N}}
\def\O{{\bf O}}
\def\P{{\bf P}}
\def\Q{{\bf Q}}
\def\R{{\bf R}}
\def\S{{\bf S}}
\def\T{{\bf T}}
\def\U{{\bf U}}
\def\V{{\bf V}}
\def\W{{\bf W}}
\def\X{{\bf X}}
\def\Y{{\bf Y}}
\def\Z{{\bf Z}}

\def\a{{\bf a}}
\def\b{{\bf b}}
\def\c{{\bf c}}
\def\d{{\bf d}}
\def\e{{\bf e}}
\def\f{{\bf f}}
\def\g{{\bf g}}
\def\h{{\bf h}}
\def\i{{\bf i}}
\def\j{{\bf j}}
\def\k{{\bf k}}
\def\l{{\bf l}}
\def\m{{\bf m}}
\def\n{{\bf n}}
\def\o{{\bf o}}
\def\p{{\bf p}}
\def\q{{\bf q}}
\def\r{{\bf r}}
\def\s{{\bf s}}
\def\t{{\bf t}}
\def\u{{\bf u}}
\def\v{{\bf v}}
\def\w{{\bf w}}
\def\x{{\bf x}}
\def\y{{\bf y}}
\def\z{{\bf z}}

\def\balpha{\mbox{\boldmath{$\alpha$}}}
\def\bbeta{\mbox{\boldmath{$\beta$}}}
\def\bdelta{\mbox{\boldmath{$\delta$}}}
\def\bgamma{\mbox{\boldmath{$\gamma$}}}
\def\blambda{\mbox{\boldmath{$\lambda$}}}
\def\bsigma{\mbox{\boldmath{$\sigma$}}}
\def\btheta{\mbox{\boldmath{$\theta$}}}
\def\bomega{\mbox{\boldmath{$\omega$}}}
\def\bxi{\mbox{\boldmath{$\xi$}}}
\def\bnu{\mbox{\boldmath{$\nu$}}}                                  
\def\bphi{\mbox{\boldmath{$\phi$}}}

\def\bDelta{\mbox{\boldmath{$\Delta$}}}
\def\bOmega{\mbox{\boldmath{$\Omega$}}}
\def\bPhi{\mbox{\boldmath{$\Phi$}}}
\def\bLambda{\mbox{\boldmath{$\Lambda$}}}
\def\bSigma{\mbox{\boldmath{$\Sigma$}}}
\def\bGamma{\mbox{\boldmath{$\Gamma$}}}

\newcommand{\myminimum}[1]{\mathop{\textrm{minimum}}_{#1}}
\newcommand{\mymaximum}[1]{\mathop{\textrm{maximum}}_{#1}}    
\newcommand{\mymin}[1]{\mathop{\textrm{minimize}}_{#1}}
\newcommand{\mymax}[1]{\mathop{\textrm{maximize}}_{#1}}
\newcommand{\mymins}[1]{\mathop{\textrm{min.}}_{#1}}
\newcommand{\mymaxs}[1]{\mathop{\textrm{max.}}_{#1}}  
\newcommand{\myargmin}[1]{\mathop{\textrm{argmin}}_{#1}} 
\newcommand{\myargmax}[1]{\mathop{\textrm{argmax}}_{#1}} 
\newcommand{\myst}{\textrm{s.t. }}

\newcommand{\denselist}{\itemsep -1pt}
\newcommand{\sparselist}{\itemsep 1pt}

\definecolor{pink}{rgb}{0.9,0.5,0.5}
\definecolor{purple}{rgb}{0.5, 0.4, 0.8}   
\definecolor{gray}{rgb}{0.3, 0.3, 0.3}
\definecolor{mygreen}{rgb}{0.2, 0.6, 0.2}

\newcommand{\cyan}[1]{\textcolor{cyan}{#1}}
\newcommand{\red}[1]{\textcolor{red}{#1}}  
\newcommand{\blue}[1]{\textcolor{blue}{#1}}
\newcommand{\magenta}[1]{\textcolor{magenta}{#1}}
\newcommand{\pink}[1]{\textcolor{pink}{#1}}
\newcommand{\green}[1]{\textcolor{green}{#1}} 
\newcommand{\gray}[1]{\textcolor{gray}{#1}}    
\newcommand{\mygreen}[1]{\textcolor{mygreen}{#1}}    
\newcommand{\purple}[1]{\textcolor{purple}{#1}}       

\definecolor{greena}{rgb}{0.4, 0.5, 0.1}
\newcommand{\greena}[1]{\textcolor{greena}{#1}}

\definecolor{bluea}{rgb}{0, 0.4, 0.6}
\newcommand{\bluea}[1]{\textcolor{bluea}{#1}}
\definecolor{reda}{rgb}{0.6, 0.2, 0.1}
\newcommand{\reda}[1]{\textcolor{reda}{#1}}

\def\changemargin#1#2{\list{}{\rightmargin#2\leftmargin#1}\item[]}
\let\endchangemargin=\endlist
                                               
\newcommand{\cm}[1]{}

%% file: LBSVM.bbl
\begin{thebibliography}{10}

\bibitem{boiman2008defense}
Boiman, O., Shechtman, E., Irani, M.:
\newblock In defense of nearest-neighbor based image classification.
\newblock In: CVPR. (2008)

\bibitem{bosch2007image}
Bosch, A., Zisserman, A., Muoz, X.:
\newblock Image classification using random forests and ferns.
\newblock In: ICCV. (2007)

\bibitem{lazebnik2006beyond}
Lazebnik, S., Schmid, C., Ponce, J.:
\newblock Beyond bags of features: Spatial pyramid matching for recognizing
  natural scene categories.
\newblock In: CVPR. (2006)

\bibitem{FLi2005}
Li, F.F., Perona, P.:
\newblock A {B}ayesian hierarchical model for learning natural scene
  categories.
\newblock In: CVPR. (2005)

\bibitem{DLowe}
Lowe, D.G.:
\newblock Distinctive image features from scale-invariant keypoints.
\newblock International Journal of Computer Vision (2004)

\bibitem{wang2010locality}
Wang, J., Yang, J., Yu, K., Lv, F., Huang, T., Gong, Y.:
\newblock Locality-constrained linear coding for image classification.
\newblock In: CVPR. (2010)

\bibitem{yang2009linear}
Yang, J., Yu, K., Gong, Y., Huang, T.:
\newblock Linear spatial pyramid matching using sparse coding for image
  classification.
\newblock In: CVPR. (2009)

\bibitem{szegedy2015going}
Szegedy, C., Liu, W., Jia, Y., Sermanet, P., Reed, S., Anguelov, D., Erhan, D.,
  Vanhoucke, V., Rabinovich, A.:
\newblock Going deeper with convolutions.
\newblock In: CVPR. (2015)

\bibitem{Brunet08}
Gouet-Brunet, V., Lameyre, B.:
\newblock Object recognition and segmentation in videos by connecting
  heterogeneous visual features.
\newblock Computer Vision and Image Understanding (2008)

\bibitem{TLee}
Lee, T., Soatto, S.:
\newblock Video-based descriptors for object recognition.
\newblock Image and Vision Computing (2011)

\bibitem{noceti2009spatio}
Noceti, N., Delponte, E., Odone, F.:
\newblock Spatio-temporal constraints for on-line 3d object recognition in
  videos.
\newblock Computer Vision and Image Understanding (2009)

\bibitem{JSivic2003}
Sivic, J., Zisserman, A.:
\newblock Video google: A text retrieval approach to object matching in videos.
\newblock In: ICCV. (2003)

\bibitem{DStavens2010}
Stavens, D., Thrun, S.:
\newblock Unsupervised learning of invariant features using video.
\newblock In CVPR (2010)

\bibitem{DTa2009}
Ta, D.N., Chen, W.C., Gelfand, N., Pulli, K.:
\newblock Surftrac: Efficient tracking and continuous object recognition using
  local feature descriptors.
\newblock In: CVPR. (2009)

\bibitem{yu2014innovative}
Yu, J., Zhang, F., Xiong, J.:
\newblock An innovative sift-based method for rigid video object recognition.
\newblock Mathematical Problems in Engineering (2014)

\bibitem{Arandjelovic05}
Arandjelovic, O., Shakhnarovich, G., Fisher, J., Cipolla, R., Darrell, T.:
\newblock Face recognition with image sets using manifold density divergence.
\newblock In: CVPR. (2005)

\bibitem{Lee03}
Lee, K.C., Ho, J., Yang, M.H., Kriegman, D.:
\newblock Video-based face recognition using probabilistic appearance
  manifolds.
\newblock In: CVPR. (2003)

\bibitem{RWang2008}
Wang, R., Shan, S., Chen, X., Gao, W.:
\newblock Manifold-manifold distance with application to face recognition based
  on image set.
\newblock In: CVPR. (2008)

\bibitem{liu2014video}
Liu, Y., Jang, Y., Woo, W., Kim, T.K.:
\newblock Video-based object recognition using novel set-of-sets
  representations.
\newblock In: CVPRW. (2014)

\bibitem{wan2015robust}
Wan, S., Aggarwal, J.:
\newblock Robust object recognition in rgb-d egocentric videos based on sparse
  affine hull kernel.
\newblock In: CVPR Workshops. (2015)

\bibitem{wolf}
Wolf, L., Shashua, A.:
\newblock Learning over sets using kernel principal angles.
\newblock Journal of Machine Learning Research (2003)

\bibitem{cherniavsky2012semi}
Cherniavsky, N., Laptev, I., Sivic, J., Zisserman, A.:
\newblock Semi-supervised learning of facial attributes in video.
\newblock In: First International Workshop on Parts and Attributes, in
  conjunction with ECCV. (2010)

\bibitem{cinbis2011unsupervised}
Cinbis, R.G., Verbeek, J., Schmid, C.:
\newblock Unsupervised metric learning for face identification in tv video.
\newblock In: ICCV. (2011)

\bibitem{cour2009learning}
Cour, T., Sapp, B., Jordan, C., Taskar, B.:
\newblock Learning from ambiguously labeled images.
\newblock In: CVPR. (2009)

\bibitem{nagendra2015video}
Nagendra, S., Baskaran, R., Abirami, S.:
\newblock Video-based face recognition and face-tracking using sparse
  representation based categorization.
\newblock Procedia Computer Science (2015)

\bibitem{karpathy2014large}
Karpathy, A., Toderici, G., Shetty, S., Leung, T., Sukthankar, R., Fei-Fei, L.:
\newblock Large-scale video classification with convolutional neural networks.
\newblock In: CVPR. (2014)

\bibitem{donahue2015long}
Donahue, J., Anne~Hendricks, L., Guadarrama, S., Rohrbach, M., Venugopalan, S.,
  Saenko, K., Darrell, T.:
\newblock Long-term recurrent convolutional networks for visual recognition and
  description.
\newblock In: CVPR. (2015)

\bibitem{tsochantaridis2005large}
Tsochantaridis, I., Joachims, T., Hofmann, T., Altun, Y., Singer, Y.:
\newblock Large margin methods for structured and interdependent output
  variables.
\newblock Journal of Machine Learning Research (2005)

\bibitem{BLi2001}
Li, B., Chellappa, R., Zheng, Q., Ser, S.Z.:
\newblock Model-based temporal object verification using video.
\newblock IEEE Tran. on Image Processing (2001)

\bibitem{ren2009egocentric}
Ren, X., Philipose, M.:
\newblock Egocentric recognition of handled objects: Benchmark and analysis.
\newblock In: CVPR Workshops. (2009)

\bibitem{ren2010figure}
Ren, X., Gu, C.:
\newblock Figure-ground segmentation improves handled object recognition in
  egocentric video.
\newblock In: CVPR. (2010)

\bibitem{liuvideo15}
Liu, Y., Kouskouridas, R., Kim, T.K.:
\newblock Video-based object recognition with weakly supervised object
  localization.
\newblock In: ACPR. (2015)

\bibitem{hare2011struck}
Hare, S., Saffari, A., Torr, P.H.:
\newblock Struck: Structured output tracking with kernels.
\newblock In: ICCV. (2011)

\bibitem{zhou2012learning}
Zhou, Q., Wang, G.:
\newblock Learning to recognize unsuccessful activities using a two-layer
  latent structural model.
\newblock In: ECCV. (2012)

\bibitem{hoai2012max}
Hoai, M., De~la Torre, F.:
\newblock Max-margin early event detectors.
\newblock In: CVPR. (2012)

\bibitem{shapovalova2012similarity}
Shapovalova, N., Vahdat, A., Cannons, K., Lan, T., Mori, G.:
\newblock Similarity constrained latent support vector machine: an application
  to weakly supervised action classification.
\newblock In: ECCV. (2012)

\bibitem{wu2012view}
Wu, X., Jia, Y.:
\newblock View-invariant action recognition using latent kernelized structural
  svm.
\newblock In: ECCV. (2012)

\bibitem{brostow2009semantic}
Brostow, G.J., Fauqueur, J., Cipolla, R.:
\newblock Semantic object classes in video: A high-definition ground truth
  database.
\newblock Pattern Recognition Letters (2009)

\bibitem{do2009large}
Do, T.M.T., Arti{\`e}res, T.:
\newblock Large margin training for hidden {M}arkov models with partially
  observed states.
\newblock In: ICML. (2009)

\bibitem{chang2011libsvm}
Chang, C.C., Lin, C.J.:
\newblock Libsvm: a library for support vector machines.
\newblock ACM Transactions on Intelligent Systems and Technology (TIST) (2011)

\end{thebibliography}
